\documentclass[lettersize,journal]{IEEEtran} 
\usepackage{amsmath,amsfonts}
\usepackage{algorithmic}
\usepackage{algorithm}
\usepackage{array}
\usepackage{amssymb}
\usepackage[caption=false,font=normalsize,labelfont=sf,textfont=sf]{subfig}
\usepackage{textcomp}
\usepackage{stfloats}
\usepackage{url}
\usepackage{verbatim}
\usepackage{graphicx}
\usepackage[table]{xcolor}
\usepackage{diagbox}
\usepackage{multirow}
\usepackage{color}
\usepackage{booktabs}
\usepackage{pifont}
\usepackage{cite}
\usepackage[numbers,sort&compress]{natbib}
\usepackage[backref,pagebackref=false]{hyperref}
\hypersetup{
hidelinks,
colorlinks=true,
linkcolor=black,
citecolor=black
}
\usepackage{float}
\usepackage{hyperref}
\hypersetup{hidelinks}
\usepackage{multirow}
\usepackage{arydshln}
\usepackage{makecell}
\usepackage{multirow}
\usepackage{graphicx}
\usepackage{arydshln}
\usepackage{amsmath}
\usepackage{amssymb}
\usepackage{mathtools}
\usepackage[utf8]{inputenc}
\usepackage{pifont}
\usepackage{booktabs}

\begin{document}

\title{Pathfinder for Low-altitude Aircraft with Binary Neural Network}

\author{Kaijie Yin$^{1,\dagger}$, Tian Gao$^{1,\dagger}$, and Hui Kong$^{1,*}$
\thanks{E-mail: { mc35211@um.edu.mo, gaotian970228@njust.edu.cn, huikong@um.edu.mo.}}
\thanks{$^{1}$ University of Macau, Macau.}
\thanks{$^{\dagger}$Equal contribution}
\thanks{* the corresponding author.}}


\maketitle

\begin{abstract}
A prior global topological map (e.g., the OpenStreetMap, OSM) can boost the performance of autonomous mapping by a ground mobile robot. However, the prior map is usually incomplete due to lacking labeling in partial paths. To solve this problem, this paper proposes an OSM maker using airborne sensors carried by low-altitude aircraft, where the core of the OSM maker is a novel efficient pathfinder approach based on LiDAR and camera data, i.e., a binary dual-stream road segmentation model. Specifically, a multi-scale feature extraction based on the UNet architecture is implemented for images and point clouds. To reduce the effect caused by the sparsity of point cloud, an attention-guided gated block is designed to integrate image and point-cloud features. 
To optimize the model for edge deployment that significantly reduces storage footprint and computational demands, we propose a binarization streamline to each model component, including a variant of vision transformer (ViT) architecture as the encoder of the image branch, and new focal and perception losses to optimize the model training. The experimental results on two datasets demonstrate that our pathfinder method achieves SOTA accuracy with high efficiency in finding paths from the low-level airborne sensors, and we can create complete OSM prior maps based on the segmented road skeletons. Code and data are available at: \href{https://github.com/IMRL/Pathfinder}{https://github.com/IMRL/Pathfinder}.
\end{abstract}

\section{INTRODUCTION}
In recent years, autonomous mapping based on ground robots has become a popular research area~\cite{gao2024activeloopclosureosmguided, qin2019autonomous}.
Existing methods can generally work well in small areas with simple topology but usually fail in large scenes~\cite{sun2023concave,das2014mapping,lau2022multi, PATEL20233837}. To solve this problem, some mapping methods guided by prior global maps have been proposed, such as based on the GPS coordinates of Open Street Maps (OSM)
~\cite{gao2024activeloopclosureosmguided, Hentschel2010AutonomousRN,munoz2022openstreetmap}. With prior global information, robots can plan an optimal path to follow and complete the mapping on a large scale.  However, prior maps are usually incomplete due to the loss of GPS signals or the lack of labeling in OSM (as shown in Fig.~\ref{fig: OSM}). Thus, robots cannot scout every corner of the scene even if they follow the planned path according to the prior map. 

This work addresses this issue by proposing an OSM maker using airborne sensors (LiDAR and camera). The integration of LiDAR and camera addresses the sparsity issue of LiDAR data while enhancing robustness to adverse lighting conditions (e.g., shadows) compared to single sensor solutions (as shown in Fig.~\ref{fig: dalvig}). The OSM maker can generate an entire topology of traversable road networks and provide the complete prior OSM map required to guide ground robots in autonomous mapping, acting as a mapping assistant for ground robots in a cooperative mapping framework~\cite{strongTogether, airGround, collab2021}. Like any real-time mapping and robotic exploration algorithms, the OSM maker should be as efficient as possible, given the low-cost and power-efficiency requirements. 
Thus, we propose an efficient binary dual-stream road segmentation model, where weights and activations are represented with only 1 bit. For 1-bit weights and activations, Xnor and PopCount operations~\cite{bnn} can replace floating-point multiplication and addition (MAC) operations, significantly reducing memory requirements and computational costs. To our best knowledge, it is the first binary dual-stream model that effectively fuses RGB images and LiDAR point-cloud data for road segmentation tasks, not limited to applications to low-altitude aircraft. The contributions of this work are summarized as follows:
\begin{figure}
\centering
\includegraphics[width=3.33in, keepaspectratio]{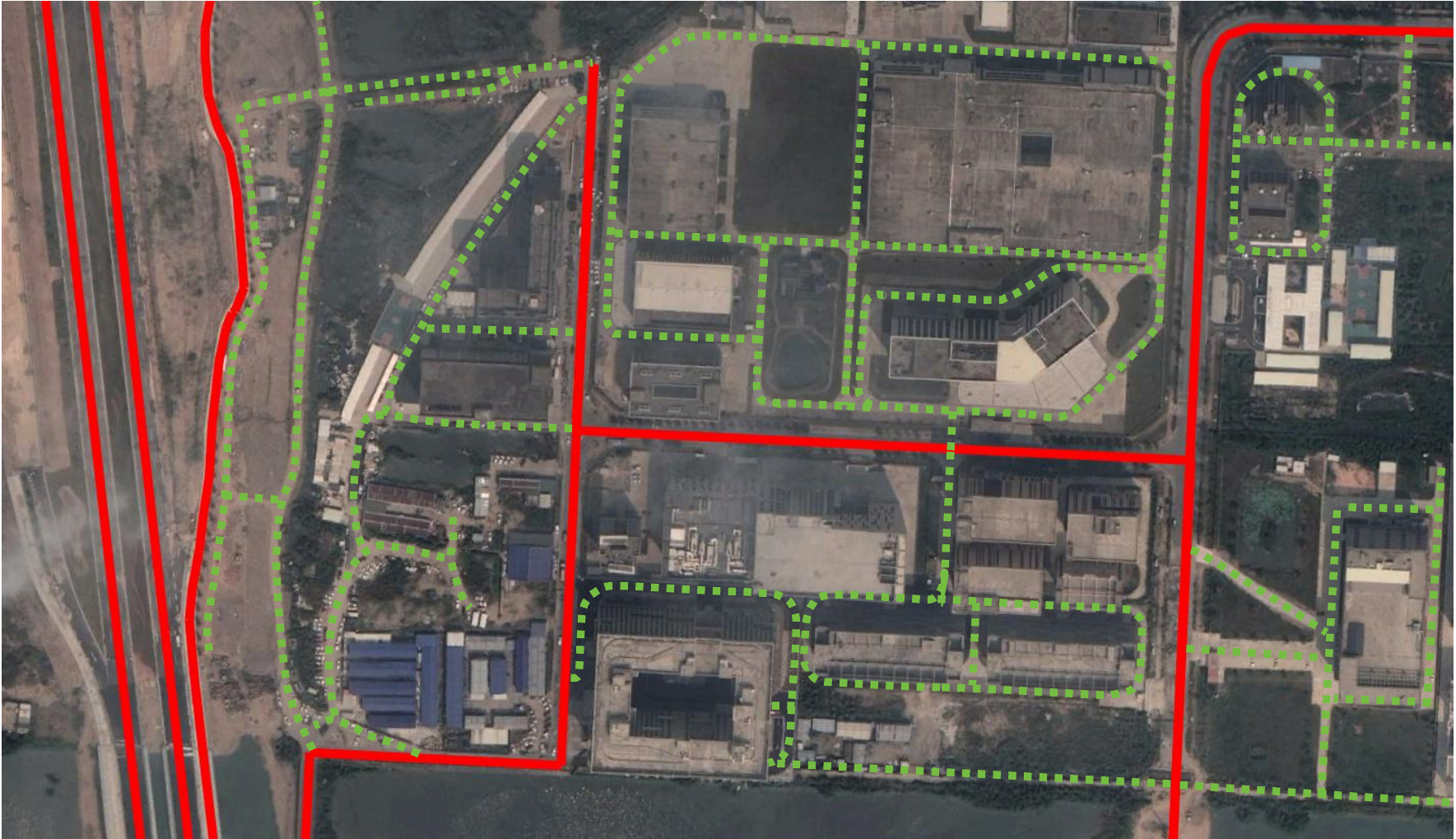}\\
\caption{OpenStreetMap Road Network on Satellite Imagery. The red lines represent the OSM roads. The areas where roads are absent from the OSM (GPS coordinates) data are marked with green dashed lines.}
\label{fig: OSM}
\end{figure}

\begin{itemize}
    \item We propose an OSM maker to automatically generate a complete OSM prior map for collaborative (ground-robot and low-altitude drone) robot mapping at a large scale. 
    \item We propose a novel binary dual-stream framework, Binarized Pathfinder, of accurate and efficient road segmentation for OSM generation. Additionally, we propose the RS-LVF dataset to support this task.
    \item To optimize the model training process, we design a new variant of focal loss to mitigate the adverse effects of class imbalance on road segmentation tasks.
\end{itemize}


\section{RELATED WORK}

\subsection{Road segmentation}
Road segmentation from aerial-view images can be categorized into traditional-based and deep learning-based. The early methods are based on manually designed features ~\cite{valero2010advanced, Yuan2013RoadSI, shao2010application}. Generally, these methods are not robust to variations in lighting and contrast~\cite{kahraman2018road}. In contrast, deep-learning methods have performed better due to their powerful representation ability. RoadNet~\cite{liu2018roadnet} was proposed to extract road surfaces, but its speed and accuracy were not optimal. To overcome this issue, FCN replaced fully connected layers with deconvolution to achieve per-pixel classification, resulting in improved performance in road segmentation tasks~\cite{kestur2018ufcn}. To effectively utilize multi-scale information, UNet was applied in road segmentation~\cite{roadDetectionUnet}. Most recently, the method based on the Deeplab network~\cite{chen2018encoder} adopted dilated convolutions and pyramid pooling layers to capture long-range information and preserve spatial structures, respectively.

Additionally, effective LiDAR-based point cloud segmentation methods have also been proposed, which can be divided into direct methods~\cite{RandLA-Net} and projection-based methods~\cite{zou2021efficienturbanscalepointclouds}. In direct methods, such as RandLA-Net~\cite{RandLA-Net}, which processes the point cloud directly without any data format conversion and applies clustering to improve the segmentation efficiency. In contrast, projection methods convert 3D point clouds into 2D images (such as bird's-eye views) to leverage efficient 2D convolutional networks for enhancing real-time processing capabilities~\cite{zou2021efficienturbanscalepointclouds}.

Image-based approaches may experience performance degradation in adverse conditions, while point clouds obtained by Lidar suffer from sparsity issues. Combining the advantages of different types of sensors has made fusion methods more promising~\cite{wang2023imbalance, PMF}. IKD-Net~\cite{wang2023imbalance} mined imbalance information across modalities to extract features from heterogeneous multimodal data comprising aerial images and LiDAR.


Due to operational efficiency and power consumption limitations, it should be noted that current methods are essentially unsuitable for edge devices such as drones. Therefore, more efficient methods such as Binary Neural Network (BNN) based methods are necessary.

\subsection{Binary Neural Network}
Network binarization~\cite{bnn, BHVIT,reactnet} is the most extreme form of network quantization~\cite{Han2015DeepCC,qin2020binary}, in which weights and activations are quantized to 1 bit~\cite{bnn}. Compared to full-precision networks, binarized models suffer from reduced accuracy due to the limited representational capacity of features and the non-differentiability of binary functions. To address this problem, various methods have been proposed to mitigate the negative impact of network binarization and improve performance without significantly increasing computational complexity. BNN \cite{bnn} is the first method that fully binarizes CNN weight and activations and applies Xnor and PopCount operations to simplify full precision matrix multiplication. Xnor-Net~\cite{xnornet} adds scaling factors to the weights and activations to enhance the performance of BNN on large datasets. By adding learnable thresholds before the sign function and incorporating RPReLU activation functions after each residual connection, ReActNet~\cite{reactnet} further improves the performance of BNN in large datasets with the reshaped output distribution. In the ViT domain, BHViT~\cite{BHVIT} demonstrates the incompatibility between the ViT structure and binarization techniques and proposes the first binarized hybrid ViT model. BiViT~\cite{bivit} proposes a mixed-precision ViT approach, where the activations in the MLP are full-precision, but the weights of all the attention and MLP modules are binary.  Binary-DAD-Net~\cite{BinaryDADnet} fully binarizes the DeeplabV3 network of full precision, marking the first BNN tailored for image-based road segmentation tasks.

\section{METHOD}
\begin{figure*}
\centering
\includegraphics[width=6in, keepaspectratio]{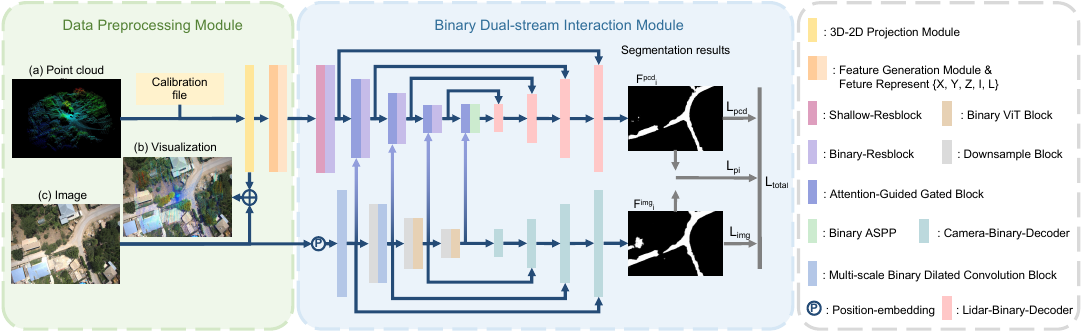}\\
\caption{The general Structure of proposed network}
\label{fig: pipeline}
\end{figure*}

The general structure of our proposed network is depicted in Fig.\ref{fig: pipeline}. Binarized Pathfinder is a typical dual-stream architecture, including an image branch and a point-cloud one. Each branch is a UNet composed of the encoder and decoder based on a pyramid structure (four stages with different feature resolutions). For each stage of the decoder, the feature from the corresponding stage of the encoder is passed through a residual link and added together. To solve the problem caused by the sparsity of the point cloud, at each stage of the encoder, we fuse image features with point cloud features using a fusion module. Additionally, we propose a new road detection dataset, the Lidar-visual-fusion dataset for road segmentation (RS-LVF), based on the MARS-LVIG dataset~\cite{lvig} for the road segmentation task. To accomplish feature alignment and the data fusion of the point cloud and image, we use the BEV projection~\cite{zou2021efficienturbanscalepointclouds} to map sparse 3D point clouds in the 2D camera coordinate system.



\subsection{Binary Convolution Unit}
\label{section BCU}
\begin{figure}
\centering
\includegraphics[width=3.33in, keepaspectratio]{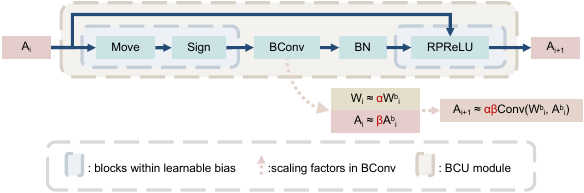}\\
\caption{The figure of the convolution process of Binary Convolution Unit (BCU)}
\label{fig: unit}
\end{figure}
\begin{figure}
\centering
\includegraphics[width=3.33in, keepaspectratio]{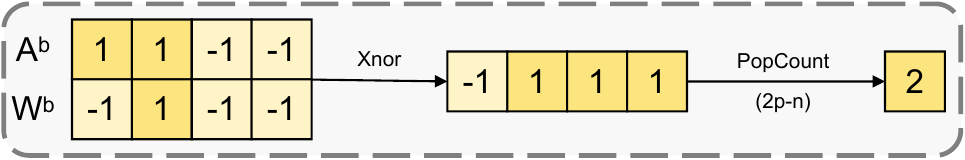}\\
\caption{The Xnor and PopCount operations.
PopCount counts the number of "1"s (the sum of 1) in the result vector of the Xnor operation (consisting of -1 and 1). $p$ and $u$ mean the number of 1 and -1, respectively. And the vector length $n=p+u$.  we can replace the dot product result from $p-u$ to $p-(n-p)=2p-n$.}
\label{fig: xnor}
\end{figure}
The Binary Convolution Unit (BCU)~\cite{reactnet} (shown in Fig.~\ref{fig: unit}) with the binary convolution layer could be represented as:
\begin{equation}
\label{con: binaryAW}
\begin{aligned}
A_{i+1}=bcov\left(\alpha W_{i}^{b},\beta A_{i}^{b} \right) \approx\alpha \beta \cdot \varphi \left(W_{i}^{b},A_{i}^{b}\right),
\end{aligned}
\end{equation}
where $W_{i}^{b}=\mathrm{sign}\left( W_i \right)$ with $W_i$ being the full-precision weight. $A_{i}^{b}=\mathrm{sign}\left( A_i \right)$ and $A_{i}$ is the full-precision activation. The $\mathrm{sign}$ is a thresholding operation. $\varphi$ is the Xnor-PopCount operation, shown in Fig.\ref{fig: xnor}, which can theoretically result in a 64 $\times$ reduction of the required inference time~\cite{xnornet} compared to matrix multiplication with full precision. $\alpha$ and $\beta$ are scale factors for weight and activation, respectively. The scale factor enables the estimated binary weights and activations to approximate the full-precision weights and activations, as indicated by $W_{i}\approx\alpha W_{i}^{b}$ and $ A_i\approx\beta A_{i}^{b}$.

To address the non-differentiability of the sign function, we apply the sub-gradient method~\cite{boyd2003subgradient}, straight-through estimator (STE), as below
\begin{equation}
\label{con: ste}
\begin{aligned}
\frac{\partial}{\partial x}\text{sign}\left( x \right) \approx \left\{ \begin{array}{l}
	1\ \ \ \ |x|<1\\
	0\ \ \ \ \text{otherwise}\\
\end{array} \right..
\end{aligned}
\end{equation}

To improve the performance of binary networks, following ReaAct-Net~\cite{reactnet}, we apply the residual connection to improve the activation information. Meanwhile, we apply learnable biases in sign and activation to mitigate gradient mismatch by adjusting the activation distribution~\cite{reactnet}.

\subsection{Binary  Encoder for Feature Extraction}
\begin{figure}
\centering
\includegraphics[width=3.33in, keepaspectratio]{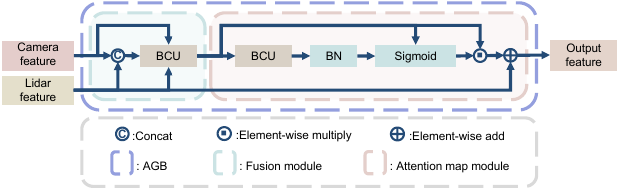}\\
\caption{Depiction of the Attention-Guided Gated Block (AGB): The AGB integrates features from both the camera and LIDAR, effectively augmenting the deficient areas in the original LIDAR feature representation. $BN$ is the batch normalization layer. $ Concat $ is the operation that concatenates two feature maps along the channel dimension.}
\label{fig: agb}
\end{figure}

\textbf{Attention-guided Gating Block (AGB):}
The sparsity of the point cloud makes the corresponding feature maps exhibit numerous voids, represented by dark hues in column 5 of Fig.~\ref{fig: dataset}. To mitigate these voids, we propose an attention-guided gating module to generate the weight to fuse image features with point cloud features, as shown in Fig.~\ref{fig: agb}. Let the image features be $ Q_i \in \mathbb{R}^{C_i~\times H_i~\times W_i} $, where $i~\in [1,2,3,4]$ is the index of the encoder stage. Similarly, let $ P_i^{2d}~\in~\mathbb{R}^{C_i~\times~H_i~\times~W_i} $ represent the features of the projected point cloud. The AGB module comprises fusion and attention map modules with the operational formula as follows
\begin{equation}
\label{con: agb}
\begin{aligned}
F_{i}^{fu}&=f_{i}^{bcu}\left( \varUpsilon \left( P_{i}^{2d},Q_i \right)+P_{i}^{2d}+Q_i \right),
\\
F_{i}^{out}&=P_{i}^{2d}+\sigma \left( F_{i}^{fu}+\eta \left( f_{i}^{bcu}\left( F_{i}^{fu} \right) \right) \right) \odot F_{i}^{fu} ,
\\
\end{aligned}
\end{equation}
where $F_{i}^{fu}$ is the output of fusion module. $f_{i}^{bcu}$ is the binary convolution unit mentioned in Section~\ref{section BCU}. $\varUpsilon $ indicates the operation that concatenates the image and lidar features along the channel dimension. $F_{i}^{out}$ means the output of the AGB block. $\sigma$ means a sigmoid function and $\eta$ represents the batch-norm layer.

\begin{figure}
\centering
\includegraphics[width=3.33in, keepaspectratio]{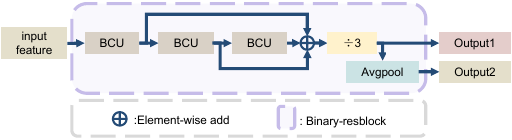}\\
\caption{Description of feature Process in a Binary-resblock}
\label{fig: resblock}
\end{figure}

\textbf{LiDAR Encoder:}
The LIDAR stream encoder comprises Binary-ResBlocks, as shown in Fig.~\ref{fig: resblock}, with four different resolutions of the feature map. The architecture of a Binary-ResBlock consists of three BCUs. The shortcut added layer by layer mitigates the gradient mismatch~\cite{reactnet} issue. The outcomes of the three BCUs are aggregated and averaged to produce output1 for residual connections to the decoder. Subsequently, an average pooling layer is applied to reduce the resolution of output1 by half as output2 for further processing. To balance performance and efficiency, we apply the shallow-ResBlock with full precision, containing only one 1x1 convolution and two convolution units, for the initial feature extraction stage of the point cloud.

\textbf{Binary Atrous Spatial Pyramid Pooling block (Binary-ASPP):}
The block consists of four binary dilated convolution units with different dilation rates ($d = \{1, 6, 12, 18\}$). These units each output feature maps of the same size. For processing these feature maps in a binarization-friendly way, different from the architecture of the original ASPP module, we replace the 1$\times$1 convolution layer with add and average operation to produce the final output.
We add the bottleneck module Binary-ASPP within the LiDAR stream, which can adjust the feature map with lowest spatial resolution obtained from the encoder to enhance feature representation.

\begin{figure}
\centering
\includegraphics[width=3.33in, keepaspectratio]{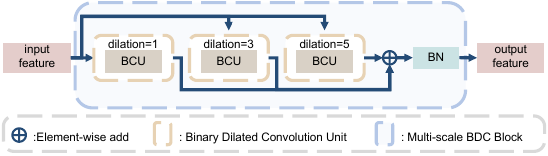}\\
\caption{The Multi-scale Binary Dilated Convolution Block(MBB) }
\label{fig: mbb}
\end{figure}

\textbf{Camera Encoder:}
The camera encoder is a multi-scale binarized CNN-ViT hybrid~\cite{bhvit}. A shallow convolutional network (with Multi-scale Binary dilated convolution Block, MBB, Fig.\ref{fig: mbb}) reduces computation, while a deeper ViT block captures rich features. This design alleviates information loss from binarizing multi-scale frameworks. The encoder uses downsampling at each layer to preserve essential details and maintain overall performance.

\subsection{Binary Decoder for Segmentation}
The LiDAR stream decoder utilizes the same binary ResBlock as the corresponding encoder. To avoid introducing additional computation, we applied pixel shuffle to perform upsampling, which only rearranges features without additional operations. Compared with bilinear upsampling, pixel shuffle is particularly suitable for BNN. For the camera stream decoder, we adopt a pyramid structure based on four binary ResNet blocks, with the resolution increasing by two times at each layer.

\subsection{Loss Function}
We employ four types of losses, including focal loss~\cite{lin2017focal}, variant focal loss, Lov\'asz loss~\cite{berman2018lovasz}, and pixel-level interaction loss, denoted by $L_{foc}$, $L_{vf}$, $L_{lo}$ and $L_{pi}$, respectively. In particular, the $L_{vf}$ is a new loss designed based on the focal loss (Eq.~\ref{con: vfloss}), through which we aim to gradually reduce the contribution of easy samples to the overall loss in a simulated annealing style. Basically, in the later training stages, the optimization focuses mainly on hard samples.
\begin{equation}
\label{con: vfloss}
\begin{aligned}
L_{vf}&=\frac{1}{C\!\!\times\!\! H\!\!\times \!\!W}\underset{c=1}{\overset{C}{\sum{}}}\underset{i=1}{\overset{H\times W}{\sum{}}}f^{mask}\!\cdot\!\hat{a}^c\!\!\cdot\!\left( -\left( 1-p^{c}_i \right) ^{\hat{\delta}}\log \left( p^{c}_i \right) \right),
\end{aligned}
\end{equation}
where $C$ is the number of classes, and $H$ and $W$ represent the height and width of the feature map, respectively. $f^{mask}$ is the void mask with 0 values corresponding to the void areas of the point cloud. The $\hat{a}^c$ and $\hat{\delta}$ are defined as
\begin{equation}
\label{con: vflossa}
\begin{aligned}
\hat{a}^c&=\left\{ \begin{array}{l}
	a_{0}^{c}\ \ \ ep\le \frac{ep_{t}}{10},\\
	a_{0}^{c}-\frac{\left( a_{0}^{c}-1 \right) \left( 10\cdot ep-ep_t \right)}{ep_t\cdot \left( \lambda -1 \right)+\varepsilon}\ \ \ \frac{ep_{t}}{10} < ep\le ep_{t}\cdot \frac{\lambda}{10}\\
	1\ \ \ \ \ \text{otherwise}\\
\end{array} \right.,
\\
\hat{\delta}&=\left\{ \begin{array}{l}
	2\ \ \ \ ep\le \frac{ep_{t}}{10}\\
	0\ \ \ \ \text{otherwise}\\
\end{array} \right.,
\end{aligned}
\end{equation}
where $\lambda$ is a hyperparameter, set 2 in the experiment. $\varepsilon$ is a small nonzero number. $ep$ is the current epoch index and $ep_{t}$ is the number of total epochs. $a_{0}^{c}=\frac{N_{total}}{N_c}$, with $N_{total}$ and $N_c$ being the number of total samples and the number of samples of class $c$, respectively.

Meanwhile, we introduce $L_{pi}$ (Eq.~\ref{con: piloss}), a pixel-level interaction loss inspired by knowledge distillation techniques, which involves calculating the Kullback-Leibler divergence of the prediction results. Specifically, we utilize the overlapping predicted results of the image feature maps and the void area of point-cloud feature maps as soft labels, which can guide the corresponding areas in the point-cloud feature maps to enhance the overall performance.
\begin{equation}
\label{con: piloss}
\begin{aligned}
& L_{pi}=\frac{1}{H\times W}\,\,\underset{i=1}{\overset{H\times W}{\sum{}}}\left( f^{mask} \right) ^{-1}KL\left( F_{i}^{pcd}||F_{i}^{img} \right),
\end{aligned}
\end{equation}
where $ KL( \cdot || \cdot )$ is the Kullback-Leibler divergence. $ F_{i}^{pcd}$ and $F_{i}^{img}$ represent the confidence map of the LiDAR stream and camera stream, respectively. Above all, the total loss is $L_{total}=L_{pi}+L_{foc}+L_{lo}^{pcd}+L_{vf}+L_{lo}^{img}$,
where $L_{lo}^{pcd}$ and $L_{lo}^{img}$ are the Lov\'asz loss applied to the LiDAR stream and camera stream, respectively.
\subsection{The post-processing of OSM maker}
\begin{figure}
\centering
\includegraphics[width=3.33in, keepaspectratio]{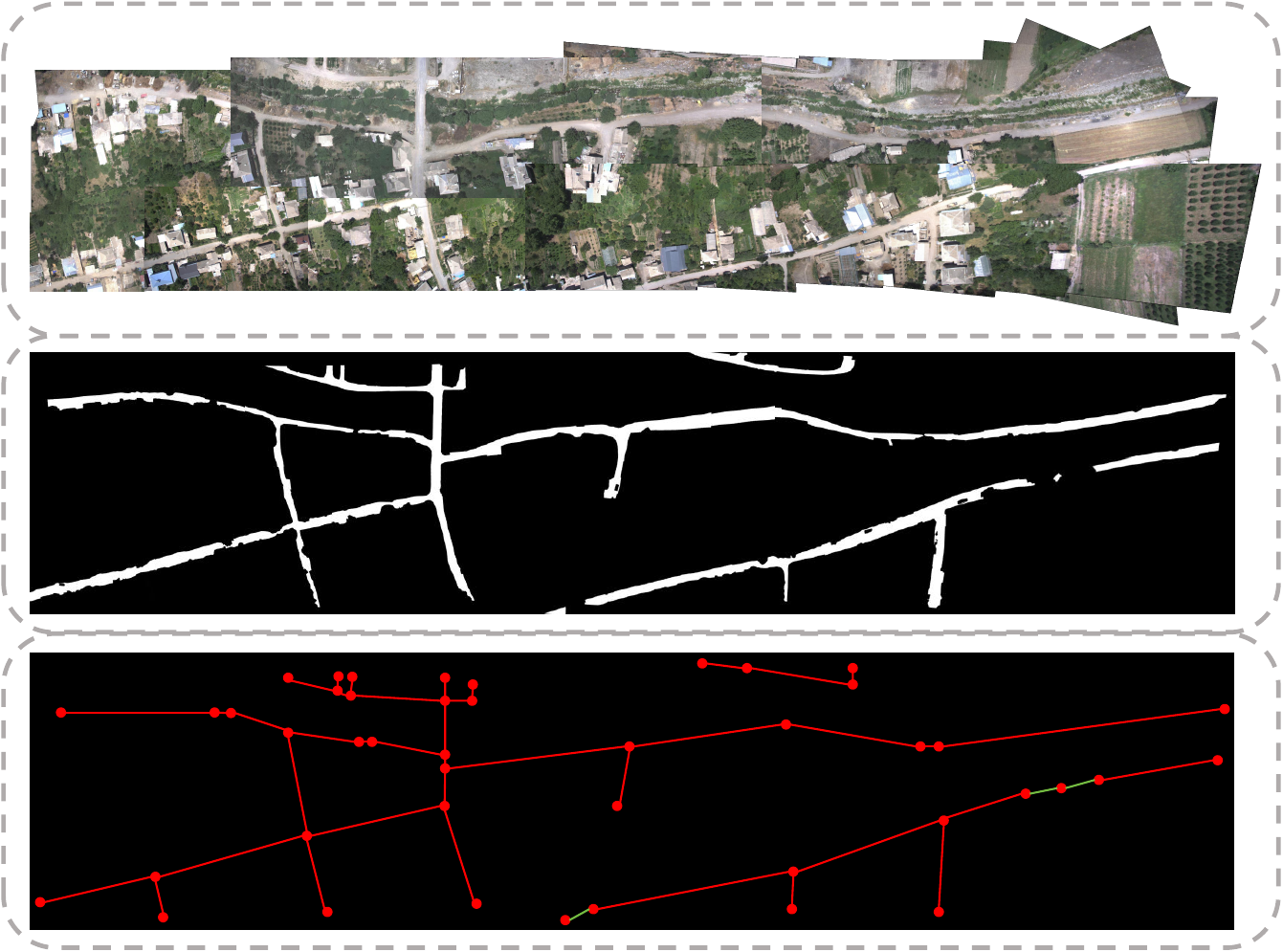}\\
\caption{The OSM maker. From top to bottom, image matching, road segmentation, and topology extraction, respectively. The green line in third row is the result of road breakpoint completion.}
\label{fig: oossmm}
\end{figure}

In this subsection, we briefly introduce the post-processing of the OSM maker, shown in Fig.\ref{fig: oossmm}. The homogeneity matrix between successive frames is estimated for image stitching. Each road segmentation map is warped according to the corresponding homography matrix to concatenate with the previous one. Then, a skeleton extraction algorithm~\cite{bulatov2017chain} is applied to extract the topological structure. In practical applications, the stitched map is aligned automatically with the existing OSM road skeleton based on a registration algorithm. Then, with interpolation of GPS coordinates, we can obtain a complete OSM map.

\section{EXPERIMENTS AND RESULTS}
\subsection{Benchmark Datasets}
\textbf{RS-LVF Dataset:}
\begin{figure}
\centering
\includegraphics[width=3.33in, keepaspectratio]{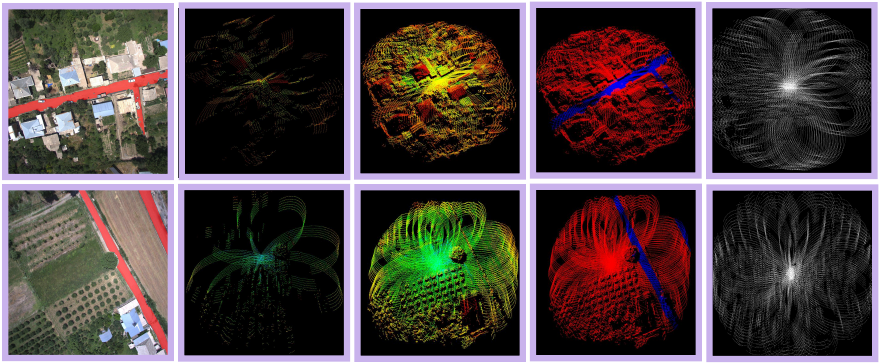}\\
\caption{The pictures from left to right are aerial images(with red labels), point clouds in a single frame, Point clouds accumulated over multi-frames(one second), point cloud labels, and point cloud projected onto image plane showing the void, respectively.
The black regions in the fifth column of the figure denote the area without point data, reflecting the typical characteristics of non-re-scanning solid-state LiDAR data. Compared with the second column, the accumulated point cloud data shown in the third column is denser.}
\label{fig: dataset}
\end{figure}
Despite numerous UAV aerial image datasets, outdoor scene datasets with multi-modal data and a downward-looking perspective are still lacking for our task. One suitable existing dataset that satisfies our requirement is the AMtown part of MARS-LVIG dataset~\cite{lvig}, whose point cloud data are collected by a non-rescanning solid-state LIDAR data with a limited FOV (e.g., the DJI-mid360~\cite{mid360}).
However, this dataset lacks semantic annotation, and each frame of the point cloud is too sparse.
Thus, we create a dataset, Lidar-visual-fusion dataset for road segmentation(RS-LVF), from the AMtown part of MARS-LVIG dataset to fulfil our road segmentation task. Totally, our dataset includes 960 pairs of LIDAR point clouds, RGB images, LiDAR-camera calibration files, and semantic annotations. Fig.\ref {fig: dataset} shows two samples. To solve the problem that the point cloud of the MARS-LVIG dataset is too sparse, we obtain frames of relatively dense point clouds by accumulating multiple sparse point cloud frames within one-second intervals based on the odometry estimated by FastLIO2~\cite{fastlio2}.




\textbf{SensatUrban Dataset:}
Generated by drone photogrammetry, SensatUrban~\cite{hu2022sensaturban} is a color-labeled point cloud dataset on the city scale. The dataset contains three billion points with rich semantic labels (categorized into 13 classes). To ensure fairness in comparisons, we follow the official partitioning strategy to train our method.
\subsection{Training Procedure and Performance Metrics}
\textbf{Training Procedure:}
For data augmentation, we apply the random color jitter to the image data. To train our model, we use the SGD optimizer with cosine annealing learning-rate decay, an initial learning rate of 0.001 and a batch size of 2 for the camera stream. Meanwhile, we utilize the AdamW optimizer with an initial learning rate set of 0.0005 for the LIDAR stream. We train our model with 100 epochs in the RS-LVF dataset. For the SensatUrban dataset, we train our model with 300 epochs.

\textbf{Performance Metrics:}
We use the Mean Intersection over Union (mIOU) metric to evaluate segmentation performance, measuring the overlap area of each class between the predicted segmentation and the ground truth: $mIoU=\frac{1}{n}\underset{c=1}{\overset{n}{\sum{}}}\frac{TP_c}{TP_c+FP_c+FN_c}$,
where $TP_c$ means true positives for class $c$, $FP_c$ and $FN_c$ mean the false positives and false negatives, respectively.
\subsection{Results}
\begin{figure*}
\centering
\includegraphics[width=6.5in, keepaspectratio]{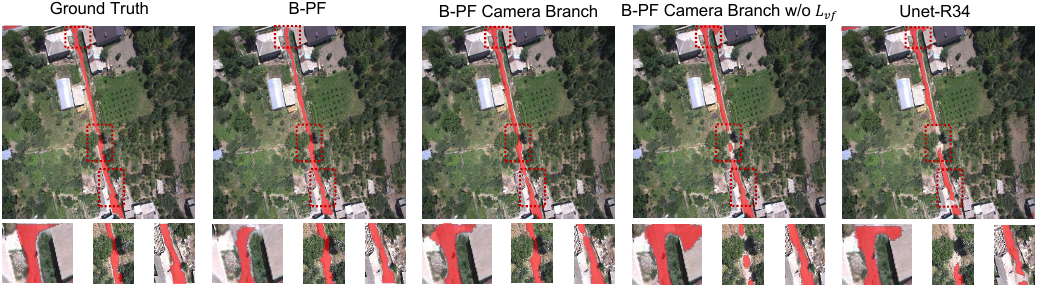}\\
\caption{Quantitative results on different scenarios in RS-LVF dataset. B-PF is an abbreviation for Binarized Pathfinder.}
\label{fig: dalvig}
\end{figure*}
\begin{table*}[htbp]
\renewcommand\arraystretch{1.1}
\centering
\caption{Comparison results of road segmentation in RS-LVF dataset. 'I' is image and 'P' refers point cloud.}
\scalebox{1.0}{
\begin{tabular}{cccccccccccc}
\Xhline{1.5pt}
\multirow{2}{*}{Method}  & \multirow{2}{*}{Input} & \multirow{2}{*}{Pre-trained} & \multicolumn{4}{c}{Point Cloud} & \multicolumn{4}{c}{Image} \\
\cmidrule(r){4-7} \cmidrule(r){8-11}
 &  &  &mAcc & mIou & Road Acc &  Road Iou& mAcc & mIou & Road Acc &  Road Iou   \\
\Xhline{1.5pt}
 SalsaNext~\cite{salsanext} & P & \ding{53} & 78.73 & 67.18 & 60.49 & 39.51 & $-$ & $-$ & $-$ & $-$    \\

 UNet R34~\cite{zou2021efficienturbanscalepointclouds} & I & $\checkmark $ & $-$ & $-$ & $-$ & $-$ & 85.36 & 77.76 & 72.06 & 58.41     \\

 PMF~\cite{PMF} & I \& P & $\checkmark $ & 95.12 & 90.47 & 91.28 & 82.19 & 81.36 & 76.54 & 63.62 & 56.54     \\

 Ours(FP) & I \& P & $\checkmark $ & \textbf{95.55} & \textbf{91.23} & \textbf{91.72} & \textbf{83.60} & \textbf{93.36} & \textbf{86.57} & \textbf{87.62} & \textbf{74.67}     \\

 Ours(binary) & I \& P & \ding{53} & 94.61 & 90.19 & 89.87 & 81.69 & 92.23 & 85.13 & 85.43 & 71.98     \\
\Xhline{1.5pt}
\end{tabular}}
\label{tab:dalvig} 
\end{table*}
\textbf{Results on RS-LVF:} As shown in Tab.~\ref{tab:dalvig}, our proposed method achieves SOTA performance among various methods trained on the RS-LVF dataset. Without fine-tuning, the proposed full-precision method achieves 91.23$\%$ point cloud mIOU and 86.57$\%$ image mIOU on the RS-LVF dataset. Compared with the corresponding full-precision network, the precision of the binary method decreases by 0.77$\%$ point cloud mIOU and 1.44$\%$ image mIOU, respectively. As shown in Fig.~\ref{fig: dalvig}, some qualitative results illustrate that integrated point cloud features have a better discriminative ability for roads with shadows or roofs with textures similar to roads.

\textbf{Results on SensatUrban:} To comprehensively evaluate the proposed algorithm, we utilize a multi-class segmentation dataset, the SensatUrban dataset, to compare the binarized pathfinder model with other SOTA methods. As shown in Table~\ref{tab:sensaturban}, the full-precision pathfinder achieves the best performance compared to the other method. Compared to the corresponding full-precision network, the accuracy of our binarized pathfinder model decreased by 2.6$\%$.

\begin{table*}[htbp]
\renewcommand\arraystretch{1.1}
\centering
\caption{Comparison results of Semantic Segmentation in SensatUrban dataset.Highlight the highest value using bold font, and the second highest using underline.}
\setlength{\tabcolsep}{1.5mm}{
\begin{tabular}{cccccccccccccccc}
\Xhline{1.5pt}
 \multirow{2}{*}{Method} & \multirow{2}{*}{mIOU}  & \multirow{2}{*}{Ground} & \multirow{2}{*}{Veg.} & \multirow{2}{*}{Build.} & \multirow{2}{*}{Wall} & \multirow{2}{*}{Bri.} & \multirow{2}{*}{Park.} & \multirow{2}{*}{Rail} & \textbf{Traffic} & \textbf{Street}& \multirow{2}{*}{Car} & \multirow{2}{*}{Foot}  & \multirow{2}{*}{Water} \\
   &   & &  & &  &  &  & & \textbf{road} & \textbf{road} &  &  &  & \\
\Xhline{1.5pt}


 RandLA-Net~\cite{RandLA-Net}  & 52.69     & 80.11 & 98.07 & 91.58 & 48.88 & 40.75 & 51.62 & 0.00 & 56.67 & 33.23 & 80.14 & 32.63  & 71.31 \\

 KPConv~\cite{kpconv}      & 57.58     & \textbf{87.10} & \textbf{98.91} & 95.33 & 74.40 & 28.69 & 41.38 & 0.00 & 55.99 & 54.43 & 85.67 & 40.39  & \textbf{86.30} \\

 UNet-R34~\cite{zou2021efficienturbanscalepointclouds}    & 59.90    & 78.45 & 95.39 & \textbf{97.89} & 60.72 & \textbf{59.80} & 40.45 & 40.35 & \textbf{67.32} & 51.42 & 82.11 & 41.94  & 62.87 \\

 MVP-Net~\cite{mvpnet}     & 59.40     & 85.10 & \underline{98.50} & 95.90 & 66.00 & 57.50 & \underline{52.70} & 0.00 & 61.90 & 49.70 & 81.80 & \textbf{43.90}  & \underline{78.20} \\

 LACV-Net~\cite{lacvnet}    & \underline{61.30}     & \underline{85.50} & 98.40 & 95.60 & 61.90 & \underline{58.60} & \textbf{64.00} & 28.50 & 62.80 & 45.40 & 81.90 & \underline{42.40}  & 67.70 \\

 NLA-GCL-Net~\cite{nlagcl} & 56.40     & $-$ & $-$ & $-$ & $-$ & $-$ & $-$ & $-$ & $-$ & $-$ & $-$ & $-$  & $-$ \\

 Ours(FP)    & \textbf{62.65}     & 82.77 & 97.70 & \underline{96.15} & \underline{74.71} & 42.96 & 40.99 & \underline{58.22} & \underline{65.50} & \underline{54.84} & \underline{85.71} & 37.43  & 77.57 \\

 Ours(binary)  & 60.03    & 82.78 & 97.31 & 95.34 & \textbf{77.44} & 11.53 & 45.65 & \textbf{62.82} & 62.32 & \textbf{55.33} & \textbf{87.18} & 35.33 & 67.32 \\
\Xhline{1.5pt}
\end{tabular}}
\label{tab:sensaturban} 
\end{table*}
\subsection{Ablation Study}
\textbf{Effect of Network Components:}
Sufficient ablation experiments are conducted to assess the effectiveness of each proposed module. We individually remove each module from the proposed binarized pathfinder model to observe performance variations on the RS-LVF dataset. The results are presented in Tab.~\ref{tab:xr}, demonstrating that each proposed module has effectively improved the network performance. The baseline represents our proposed algorithm stripped of the mentioned modules. $mIOU^{*}$ is the average mIOU of the point cloud and the image.
Specifically, the difference of $mIOU^{*}$ between the fifth row and the last row of Tab.~\ref{tab:xr} illustrates the effectiveness of the proposed loss $L_{vf}$. The quantitative result (shown in Fig.~\ref{fig: dalvig}) demonstrates that the proposed loss can recognize hard samples.

\begin{table}[htbp]
\renewcommand\arraystretch{1.1}
\centering
\caption{Performance metrics for different models and classes. "Bi-ASPP" refers to utilizing the optimized version of ASPP to instead the original binary ASPP. }
\setlength{\tabcolsep}{5pt}{
\begin{tabular}{cccccccc}
\Xhline{1.5pt}
 \multirow{2}{*}{Baseline} &\multirow{2}{*}{AGB} & \multirow{2}{*}{Bi-ASPP} & \multirow{2}{*}{$L_{Pi}$} & \multirow{2}{*}{$L_{vf}$} &  \multicolumn{3}{c}{$mIOU$}\\
 \cmidrule{6-8}
  $ $ &  &  &  & & LiDAR&Img& All\\
\Xhline{1.5pt}
 $\checkmark $ &  &  &  & & 62.65 & 83.81 &73.2 \\

 $\checkmark $ & $\checkmark $ &  &  &  &87.83&84.98& 86.4 \\

 $\checkmark $ &  & $\checkmark $ &  & & 63.31&83.77&73.6\\

 $\checkmark $ & $\checkmark $&  & $\checkmark $ & & 88.79 & 85.04 &86.9 \\

 $\checkmark $ & $\checkmark $ & $\checkmark $ & $\checkmark $ &  &89.46 &  84.84&87.2 \\

 $\checkmark $ & $\checkmark $ & $\checkmark $ &  & $\checkmark $ & 89.18 & 85.10 &87.1\\

 $\checkmark $ & $\checkmark $ & $\checkmark $ & $\checkmark $ & $\checkmark $ & 90.19 & 85.13 & 87.8 \\
\Xhline{1.5pt}
\end{tabular}}
\label{tab:xr} 
\end{table}
\begin{table}[htbp]
\renewcommand\arraystretch{1.1}
\centering
\caption{Computational complexity for different models. G is $10^9$ and the image size is 512 $\times$ 512}
\begin{tabular}{cccccc}
\Xhline{1.5pt}
 Network & Ops (G) & Para. (MB) & Latency (ms) \\ 
 \Xhline{1.5pt}
 PMF-Net (FP) & 237.71 & 138.90 &  1787  \\
 Ours (FP) & 202.28 & 153.69    &  1754  \\
 Ours (binary) & 16.41 & 15.24  &  604  \\
 \Xhline{1.5pt}
\end{tabular}
 \label{tab:op} 
\end{table}
\textbf{Effect of binarization:}
Following ReAct-Net~\cite{reactnet}, we utilize the OPs ($OPs = BOPs / 64 + FLOPs$) to evaluate the complexity of binary neural networks. BOPs and FLOPs represent binary operations and floating-point operations, respectively. As shown in the Tab.~\ref{tab:op}, the binarization process decreases the OPs of Pathfinder from 202.28 G to 16.41 G. The memory requirement is also decreased from 153.69 MB to 15.24 MB.

\textbf{Latency result on edge devices.}
To measure real-time hardware latency, we first convert the PyTorch implementation of the proposed framework to ONNX format and then deploy it on an ARM Cortex-A76-based edge device (without CUDA) using the Bolt toolkit. Experimental results in Table~\ref{tab:op} demonstrate that the 1-bit version reduces latency by 2.9× compared to full-precision methods. We anticipate further acceleration potential through dedicated hardware accelerators optimized for binarized operations.


\section{CONCLUSIONS}
In this paper, we propose an OSM maker to address missing partial path labels in existing OSM methods. The core innovation is a UNet-based binary dual-stream road segmentation model, enhanced with optimized loss functions and specialized modules. Experiments demonstrate state-of-the-art mIOU with low memory/computational costs, enabling efficient high-precision OSM map completion and maintenance.

\section*{ACKNOWLEDGEMENTS}
This work was supported by the Fundo para o Desenvolvimento das Ciências e da Tecnologia of Macau (FDCT) with Reference No. 0067/2023/AFJ, No. 0117/2024/RIB2.

\end{document}